\documentclass[sigconf]{acmart}

\usepackage{hyperref}
\usepackage{subcaption}
\usepackage{soul}
\usepackage{booktabs}
\usepackage{filecontents}
\usepackage{siunitx}
\usepackage{enumitem}
\usepackage{comment}
\usepackage{xcolor}

\setlist[itemize]{noitemsep, topsep=0pt}

\AtBeginDocument{%
  }

\copyrightyear{2023}
\acmYear{2023}
\setcopyright{acmlicensed}\acmConference[GoodIT '23]{ACM International Conference on Information Technology for Social Good}{September 6--8, 2023}{Lisbon, Portugal}
\acmBooktitle{ACM International Conference on Information Technology for Social Good (GoodIT '23), September 6--8, 2023, Lisbon, Portugal}
\acmPrice{15.00}
\acmDOI{10.1145/3582515.3609515}
\acmISBN{979-8-4007-0116-0/23/09}

\title{Leave no Place Behind: Improved Geolocation in Humanitarian Documents}

\author{Enrico M. Belliardo}
\email{enrico.m.belliardo@gmail.com}
\orcid{0009-0006-4456-3384}
\affiliation{
    \institution{ISI Foundation}
    \city{Turin}
    \country{Italy}
    }
\author{Kyriaki Kalimeri}
\email{kyriaki.kalimeri@isi.it}
\affiliation{
    \institution{ISI Foundation}
    \city{Turin}
    \country{Italy}
    }
\author{Yelena Mejova}
\email{yelenamejova@acm.org}
\affiliation{
    \institution{ISI Foundation}
    \city{Turin}
    \country{Italy}
    }

\renewcommand{\shortauthors}{Belliardo et al.}

\begin{abstract}

  Geographical location is a crucial element of humanitarian response, outlining vulnerable populations, ongoing events, and available resources.
  Latest developments in Natural Language Processing may help in extracting vital information from the deluge of reports and documents produced by the humanitarian sector.
  However, the performance and biases of existing state-of-the-art information extraction tools are unknown.
  In this work, we develop annotated resources to fine-tune the popular Named Entity Recognition (NER) tools Spacy and roBERTa to perform geotagging of humanitarian texts. 
  We then propose a geocoding method FeatureRank which links the candidate locations to the GeoNames database.
  We find that not only does the humanitarian-domain data improves the performance of the classifiers (up to F1 = 0.92), but it also alleviates some of the bias of the existing tools, which erroneously favor locations in the Western countries.
  Thus, we conclude that more resources from non-Western documents are necessary to ensure that off-the-shelf NER systems are suitable for the deployment in the humanitarian sector.

\end{abstract}

\ccsdesc[500]{Social and professional topics~Geographic characteristics}
\ccsdesc[300]{Information systems~Data mining}
\ccsdesc[300]{Information systems~Information systems applications}

\keywords{Geolocation, humanitarian, location bias, geotagging, geographic bias, named entity recognition, geocoding}

\begin{document}

\maketitle

\section{Introduction}

The vast network of humanitarian organizations performing a myriad of tasks around the world produces a significant amount of data and reports. 
For instance, the operations of the International Federation of Red Cross and Red Crescent Societies (IFRC) spans 192 countries and includes almost 14 million volunteers \cite{redcross2022what}.
To manage the resulting reports, projects such as the Data Entry and Exploration Platform (DEEP)\footnote{https://thedeep.io/} have been developed to help humanitarian organizations to compile and store documentation, and to help structure the qualitative information inside.
It is maintained by Data Friendly Space, a U.S.-based international non-governmental organization (INGO)\footnote{https://datafriendlyspace.org/}, in collaboration with numerous international organizations.

In the light of the information overload generated by the data-rich operation of humanitarian efforts, computer-aided information extraction promises to increase the usefulness of the produced documentation. 
Latest developments in Deep Learning (DL) and Natural Language Processing (NLP) allow for the identification of potentially important information in text, and its classification in standard hierarchies, which can then be used to aggregate knowledge and share experience across time and projects.
In particular, geolocation is an important aspect of humanitarian activity, spanning both operations in entire countries and regions, to specific villages, refugee camps, and civil infrastructure.
The accuracy of location information is especially important in the light of the Sustainable Development Goals (SDGs) principle of ``Leave No One Behind'' that strives to reduce inequalities and vulnerabilities amongst all countries. 
Unfortunately, the data sources available for training competent models are often biased towards the West, even when secondary data is used to fill the ``datasphere'' gaps \cite{dalton2016critical}. 
Location databases (or ``gazetteers'') have a well-known bias towards the US and other Western countries \cite{graham2015hidden,graham2015mapping}, whereas popular alternative data sources such as Twitter and Wikipedia struggle with penetration in global South \cite{statista2022twitter,wikipedia2023list}.
Thus, it is imperative to create resources and tools for the accurate processing of diverse humanitarian data in order to ensure no country is disadvantaged when the information is collated.

In this study, we work closely with humanitarian partners to create a Geolocation Extraction tool tuned specifically to process documents generated by humanitarian projects. 
The tool performs two sub-tasks: \textbf{geotagging} -- the extraction of text fragments that may be a location (or ``toponyms''), and \textbf{geocoding} -- the disambiguation of the toponym to a specific geographic location (with accompanying information on GPS coordinates, type of location, etc.). 
We contribute two datasets -- one for each of these steps -- wherein humanitarian reports are annotated by specialists in the field for candidate toponyms, and these toponyms are mapped to the appropriate entry in GeoNames, an extensive geographical database\footnote{https://www.geonames.org/}.
Using these datasets, we tune popular existing Named Entity Extraction (NER) models Spacy\footnote{https://spacy.io/models/en\#en\_core\_web\_md} and roBERTa\footnote{https://huggingface.co/Davlan/xlm-roberta-base-wikiann-ner}, improving their performance up to F1 $= 0.92$ on the target dataset.
We then propose a geocoding algorithm FeatureRank tailored to the humanitarian domain, and compare it to two baselines from the literature. 
Finally, we show that the tuned model not only improves in accuracy, but that it may alleviate the geographic bias that favors locations in the Western countries.

\section{Related Literature}

% Generally about NER
Named Entity Recognition (NER) is one of the basic Natural Language Processing (NLP) tasks involving the identification of entities of interest in texts, commonly constrained to Person, Organisation or Location. 
Whereas the early models used classic machine learning algorithms including support vector machines, random forests, and decision trees using as features capitalization, word endings, and regular expressions \cite{nadeau2007survey}, from about 2011 neural networks have been used to create more generalizable models \cite{collobert2011natural,yadav2018survey}.
Combining word context and its individual characters has allowed the training of systems that required little domain-specific training data.
Word and sentence level embeddings have been used to build models using several long short-term memory (LSTM) layers to achieve performance of up to 90\% F score in English \cite{dernoncourt2017neuroner} (this model was also published as a NeuroNER package\footnote{http://neuroner.com/}).
Graph embedding algorithms that produce embeddings of vertices that preserve their proximity in a graph, including DeepWalk \cite{perozzi2014deepwalk} and TransE \cite{bordes2013translating}, have been widely used in entity linking (EL) task, the task of matching a piece of text (a potential entity) to an entry in a knowledge base such as Freebase, DBpedia, and Wikidata.
Since 2015, large pre-trained models like Bidirectional Encoder Representations from Transformers (BERT) \cite{kenton2019bert} have given researchers the access to representations trained on huge amounts of text data without the need to access this data directly, while allowing strong expressive power suitable to many NLP tasks.
These models can then be trained using the distant labels, improving task-specific recall and precision, and further enhanced using self-training \cite{liang2020bond}.
Neural networks and pre-trained large models have also been used for entity linking \cite{sevgili2022neural}.
For instance, BERT-based cross-encoders have been used on a concatenation of text segment context with a candidate entity description to produce a score for each entity candidate \cite{logeswaran2019zero,wu2020scalable}.
However, few of these efforts have been directed specifically to geographical NER.

% Geographical entity extraction for humanitarian applications
Natural language processing approaches have been applied to understand the context in the extraction and geocoding of historical floods, storms, and adaptation measures \cite{lai2022natural}. 
In the context of mass emergencies,  Imran et al.~\cite{imran2015processing} highlighted the existing approaches and challenges in processing social media messages during mass emergencies. This survey sheds light on the importance of real-time analysis of social media data for effective emergency response.
Several techniques were explored to extract and refine location mentions in text such as geoparsing, location disambiguation, and geotagging \cite{middleton2018location}.
For instance, geotagging text data on the web has been approached through geometrical methods \cite{radke2018geotagging}, offering an alternative to relying solely on explicit geotags.
The potential of Twitter data were highlighted both in determining the geographic origin of user-generated content \cite{graham2014world}, and developing predictive models that can estimate the location of Twitter users based on their posted content \cite{zheng2018survey}. 
Moreover, mining Twitter data offered a better understanding of disaster resilience \cite{zou2018mining}, while was also proven effective in event classification and location prediction during disasters \cite{singh2019event}.
Importantly, efforts have been made to develop annotated datasets and resources \cite{poletto2021developing, imran2022tbcov} contributing to the development and evaluation of algorithms for the identification and monitoring of internal displacement, but also sentiment analysis, named entity recognition, and geolocation from textual data.
In this work, we contribute annotated resources concerning the humanitarian reports, instead of general web data.

Despite these advancements there are still many open issues to accurate named entity recognition for the humanitarian sector. 
For instance, known biases of the geographic information and modes of communication are directly impacting the representation of populations unevenly~\cite{graham2015mapping}.
Population bias in geotagged on social media has been a topic of concern in geolocation research \cite{malik2015population,rama2020facebook}, highlighting the potential biases that can arise from relying solely on geotagged tweets for location inference. 
Understanding and addressing these biases is crucial to ensure accurate geolocation results.

\section{Data collection}

We employ data from the {\tt HumSet} database \cite{fekih-etal-2022-humset}, originated from the multi-organizational platform (DEEP) introduced earlier.
Each entry consists of a relevant excerpt from a document, annotated with the humanitarian analysis framework categories. It is associated to a \emph{lead\_id} that identifies the original document (\emph{lead}) from which the excerpt is extracted.
Original documents come from different sources, including reports by humanitarian organizations and media articles. We downloaded them using \emph{deepex} package\footnote{available at \url{https://github.com/the-deep/deepex}} from the links available in {\tt HumSet}. The documents come either in PDF format or HTML pages, and text extraction from both sources is performed with the same package. This parser also splits text in pages and paragraphs, according to the basic structure of the document. 
In this project, we use the entire text parsed from {\tt HumSet} \emph{leads}, and not only the selection of excerpts of the database.

\subsection{Data description}

The dataset contains \num{15661} documents from 45 different projects. It is multilingual, with English (67.1\%), Spanish (18.6\%) and French (14.1\%) representing almost the totality of the documents' languages (language information is available in {\tt HumSet}).
Each project typically focuses on one or more countries, but some focus on none. There are 33 countries represented by country-specific projects. 
The content of the documents may vary from text to images and tables. The parser is able to fetch plain text and discard figures and non textual elements. Some documents contain mainly non textual elements, therefore the resulting extracted text is very short. Table \ref{table: doc_stats} shows a distribution of metrics of the text content extracted with the parser.

\begin{table}[!htb]
    \centering
    \caption[]{Summary statistics on text files parsed from documents.}
    \label{table: doc_stats}
    \begin{tabular}{rrrrr}
        \toprule
            &size(KB)&pages&paragraphs&words\\
        \midrule
        mean&11.6    &5.0  &38.0      &1621\\
        std &40.5    &14.5 &130.9     &5487\\
        min &0.0     &1    &0         &0\\
        25\%&1.2     &1    &2         &170\\
        50\%&3.2     &1    &8         &459\\
        75\%&7.6     &3    &26        &1085\\
        max &1354.2  &358  &4254      &\num{207030}\\  
        \bottomrule
    \end{tabular}
\end{table} 

\section{Data Annotation}

This work contributes two annotated datasets, one for each step of the geolocation extraction process: (1) \textbf{geotagging}, which is a special case of NER, and (2) \textbf{geocoding}, where the found toponyms are disambiguated and linked to specific geographic coordinates \cite{gritta-etal-2019-geoparsing}.
For both tasks, we describe how the data was sampled, baseline models were applied to extract candidates, how these candidates were annotated, guided by coding schemas. Finally, we present the statistics about the resulting annotated datasets.

\subsection{Annotation: geotagging}

\subsubsection{Sampling}

We begin by selecting 500 English-language documents from {\tt HumSet}, satisfying several metrics, as described below. As we want our sample to include as many different locations as possible, while respecting the country distribution ranking of the dataset, we treat the documents belonging to the 6 most popular project countries
differently from the rest (the ``tail'' of the distribution).
We impose different restrictions on the popular and tail countries in terms of on the number of pages (minimum of 3 or 2, respectively), average number of paragraphs per page (minimum 5 or 3) and average number of words per paragraph (minimum 500 or 300).  
This filter results in \num{1897} documents. We further sample up to 100 
documents from each project country, to undersample popular countries, resulting in 1290 documents, which we then sample randomly to get the final 500 document sample. 

\subsubsection{Pre-annotation using baseline models}
\label{sec:baseline_models_geotagging}

Labelling task is performed with Label Studio app\footnote{\url{https://labelstud.io/}}. With it, is possible to upload pre-annotated documents, reducing the effort of the user to a simple revision of the pre-annotations. 
In order to insure a variety of locations in our dataset, we truncate the documents at 4000 characters, thus encouraging the annotators to label more distinct documents.
To further help annotators find the candidate toponyms, we considered pre-annotations which may ease the annotation process, letting the user to focus on difficult cases without the need to tag all the locations in the text. 
To explore this, we run a small test experiment prior the full annotation process involving 4 users and 25 pilot documents. Two users were asked to tag pre-annotated documents, the other 2 -- documents without pre-annotations. Annotation agreement does not vary among users with pre-annotated documents and users without pre-annotated documents (remaining at around 0.90), however the speed of their work improves by 7\% (and this was during an adjustment period where some aspects of the process were unclear to the labelers).
Thus, we proceed to create candidate toponyms using baseline NER models.

Pre-annotations are provided by two off-the-shelf NER models: Spacy \emph{en\_core\_web\_md}\footnote{https://spacy.io/models/en\#en\_core\_web\_md}, a pre trained pipeline for English that includes NER components; and the roBERTa \emph{xlm-roberta-base-wikiann-ner}\footnote{https://huggingface.co/Davlan/xlm-roberta-base-wikiann-ner}, a multilingual roBERTa based NER model finetuned on 20 annotated Wikipedia datasets\footnote{https://huggingface.co/datasets/wikiann}. 
The pre-annotations are the results of the union of the predictions of the 2 models. In particular, in case of multi-word entities, if the predictions of the 2 models overlap, the union of words of the 2 models is used as pre-annotation (for instance, model 1 predicts \emph{"the Mediterranean"} and model 2 \emph{"Mediterranean sea"}, the resulting union is \emph{"the Mediterranean sea"}).

\subsubsection{Annotation schema}

Defining clearly the meaning of location is crucial for the annotation task. As shown in many other works, a location-related term could take different meaning depending on the context \cite{sekine-etal-2002-extended, leidner-2008}. 
We follow the main distinction by Gritta et al.~\cite{gritta-etal-2019-geoparsing} between \emph{literal} and \emph{associative} toponyms. 
A \emph{literal} toponym refers directly to a physical location and an \emph{associative} toponym is only associated with a place. 
For example, ``Syria'' is a literal toponym in ``latest events in central Syria'', but an associative in ``Syria Red Cross aided border regions''.
The annotation task, then, comprises of reading the text, and annotating candidate geotags, as well as creating new ones, and labeling them as either of these 2 categories: literal or associative.
Instructions also include rules on handling lists of locations, which are split in post-processing, as well as how to annotate humanitarian-specific text such as names of reports, organizations, and other adjectives that may include toponyms. The instructions are available at \url{shorturl.at/cxJN1}. 

\subsubsection{Annotated geotagging dataset}

A total of four 469 documents were annotated by annotators associated with the DEEP platform, specializing in humanitarian data analysis. 
The median number of toponyms per document is \num{25}, and the total number of toponyms annotated is \num{11025}. 
Recall that each document has been truncated at 4000 characters to allow a broader inclusion of different documents.
The most common toponyms in the data were \emph{Libya, Syria, Tripoli, Afghanistan, Yemen, Niger, Sudan, Venezuela, Nigeria, Somalia}. 
The annotated dataset includes \emph{lead\_id} of the document, \emph{source} url link, \emph{text} of the document (first 4000 characters) and \emph{annotations} and is available at \url{https://github.com/embelliardo/HumSet_geolocation_annotations}.

\subsection{Annotation: geocoding}
\label{annotation:geocoding}

The second annotated dataset concerns the second task of this study: the association of toponyms to a disambiguated location, including the GPS coordinates. 
For this purpose, we have created a dataset of toponyms in context mapped to an ID of GeoNames entry. 
GeoNames is a geographical database that contains over 25 million geographical names, of which 4.8 million are populated places, which also come with 13 million alternate names in local alphabets\footnote{\url{https://www.geonames.org/about.html}}.

% Sampling / data selection
The toponyms were selected from a random selection of documents annotated in the previous section.
To prepare the toponyms for annotation, regular expressions were used to split the lists of toponyms into individual locations (i.e. ``Al Hudaydah and Taizz governorates'' to ``Al Hudaydah governorate'' and ``Taizz governorate''), associate noun modifiers to each location (i.e. ``North and South Italy'' to ``North Italy'' and ``South Italy''), rephrase possessive pronouns (i.e. ``City of New York' to ``New York City''), standardize special characters, and remove dashes and apostrophes.
% Initial matching
These cleaned toponyms were then matched with the database of GeoNames fields of \emph{name} and \emph{alternate name} (which provides additional ways of writing the location name), which were similarly preprocessed by standardizing special characters, and removing dashes and apostrophes.
The match was done in two ways.
First match was performed using longest consecutive word matching.
The matches were then filtered through a custom list which includes cardinals alone, common words, and generic location signifiers. 
Second match was performed on the same GeoNames fields using the Whoosh search engine that uses the Okapi BM25F ranking function to retrieve strings\footnote{\url{https://whoosh.readthedocs.io/en/latest/intro.html}}. 
This way of matching allowed for more flexible positioning of keywords (i.e. ``Tripoli Airport'' can be matched to ``Tripoli International Airport'', which consecutive word matching would miss).
The search engine matches were then filtered based on the ``feature code'' of the GeoNames entity, which describes what kind of location it is. 
We keep locations having this code among the following (hand-crafted based on relevance): administrative division (AD), populated place (PPL), mountain (MT), sea (SEA), lake (LK), island (ISL) and airport (AIR). 
Note that we use the same pipeline for the preparation of the toponyms for geocoding using our automated algorithm below.

% Annotation procedure
The annotation was performed using a custom-built tool that displays the toponym, the document as its context, and a list of candidate GeoNames entities, which are located on a map.
The annotators were instructed to select a GeoNames entity most likely to be referenced by the toponym, and if the correct one cannot be found, to select ``none''.
The main author of this paper performed the annotation, resulting in 561 unique document/toponym match pairs from 39 documents, with 474 having non-empty matches, spanning 78 countries.
The annotated dataset includes \emph{lead\_id} of the document, \emph{toponym}, \emph{match} and the \emph{GeoNames ID} of the correct match, and is available at \url{https://github.com/embelliardo/HumSet_geolocation_annotations}.

\section{Customizing Geo-Location for Humanitarian Texts}

In the next sections, we describe the evaluation of state-of-the-art geotagging methods and tuning of these methods with the annotated data. 

\subsection{Geotagging}

Using the annotated geotagging dataset, we were now able to evaluate the off-the-shelf NER models -- Spacy and roBERTa -- used in pre-selection of the toponyms. 
We compute two versions of the metrics: exact agreement considers only exact string matches to the ground-truth labels (subscripts $e$ in Table \ref{tab:geotagging_performance}), whereas partial agreement considers any substring overlap (subscript $p$). 
Note that partial scoring includes exact matches.
We define agreement between the output of the model and the ground-truth as the number of toponyms predicted by the model which are in the annotated set, divided by the union of toponyms of the two sets. 
We further use the standard definitions of precision and recall.
In the case of multiple correct matches, we count such matches as one (by weighting each match proportional to the number of sub-matches), such that we do not inflate the number of matched toponyms. 
Table \ref{tab:geotagging_performance} reports the performance of the baseline models (as well as the tuned ones, described below).
Both algorithms were evaluated on the entire annotated set using 10-fold cross-validation. 
The cross-validation is run in two ways: without stratification, and with stratification by the country of the project, such that the training data does not have information about the test project country, making it a more difficult task.

\begin{table*}[!htb]
    \centering
    \caption{Agreement, precision, recall, and F1 for geotagging methods: mean (standard deviation) over 10 folds. Exact matches with subscript $e$ and partial with subscript $p$. Tested with and without stratification (``ST'') by country of the project.}
    \setlength{\tabcolsep}{3pt}
    \label{tab:geotagging_performance}
    \begin{tabular}{l|rrrr|rrrr}
        \toprule
        Method & A$_e$ & P$_e$ & R$_e$ & F1$_e$  & A$_p$ & P$_p$ & R$_p$ & F1$_p$ \\
        \midrule
        roBERTa$_{baseline}$ & .63(.03) &	.73(.02) &	.72(.03) &	.72(.02) & .74(.03) &	.85(.02) &	.85(.03) & .85(.02) \\
        roBERTa$_{tuned}$ & \textbf{.75(.03)} & .78(.02) & \textbf{.81(.02)} & \textbf{.79(.02)} & \textbf{ .87(.03)} & .92(.02) &	.95(.01) & \textbf{ .93(.01)}\\
        roBERTa$_{tuned\_ST}$ & .72(.03) & .79(.04) & .76(.03) & .78(.03) & .85(.04) &	\textbf{ .94(.02)} &	.90(.04) &	.92(.02) \\
        Spacy$_{baseline}$ \rule{0pt}{2.8ex}& .53(.08) & .74(.07) & .57(.08) & .64(.07) & .64(.07) & .90(.04) & .68(.08) & .78(.05) \\
        Spacy$_{tuned}$ & .74(.03) &\textbf{.79(.03)} & .80(.03) & \textbf{.79(.02)} & .85(.02) & .92(.03) & .92(.03) & .91(.01) \\
        Spacy$_{tuned\_ST}$ & .65(.06) & .77(.06) & .70(.07) & .74(.05) & .77(.06) &	.92(.04) &	.82(.07) &	.87(.04) \\
        Combined \rule{0pt}{2.8ex} & .72(.03) & .74(.04) & .80(.02)	& .78(.03) & \textbf{ .87(.03)} & .89(.03) & \textbf{ .97(.00)} & .93(.02) \\
        Combined$_{\_ST}$ & .70(.05) & .73(.05)&	.77(.03) & .75(.04) & .86(.04) & .90(.04) & .95(.01) & .92(.03) \\
        \bottomrule
    \end{tabular}
\end{table*}

For both algorithms, and both evaluation methods, we find the tuning of the model with additional data to improve performance, both in terms of precision and recall. 
For roBERTa, the improvement in F1 is from 0.72 to 0.79 (not stratified) for exact matches and from 0.85 to 0.93 (not stratified) for partial ones. 
The improvement is even more for Spacy, a 15 percentage point improvement in F1 to 0.79 (not stratified) for exact matches, and a 13 point improvement in F1 to 0.91 (not stratified) for partial. 
Interestingly, the variance of performance of the Spacy model decreases with additional tuning, however in the stratified testing it remains higher.
As expected, the performance is lower when the data is stratified by project country, but there are still some improvements.
We also combine the two models by taking a union of the matches produced by both algorithms (and a largest span of those that partially overlap).
This approach favors the recall, achieving 0.97 in partial matching, while still having a precision of 0.89.
The training brings the performance of the two algorithms closer (with Spacy having largest improvement), but other considerations may be important in the deployment of the models. 
Whereas Spacy model takes about 6.5MB, roBERTa is much larger at 1.1GB. 
Similarly, Spacy runs faster, about 3.2 times faster than roBERTa.
To conclude, if one favors recall, a combined model should be used; if a low variance in performance is desired, roBERTa may be a better choice; but if a lightweight model is needed for fast processing, Spacy may be a better selection.

\begin{figure*}[!htbp]
    \begin{center}
        \includegraphics[width=13cm]{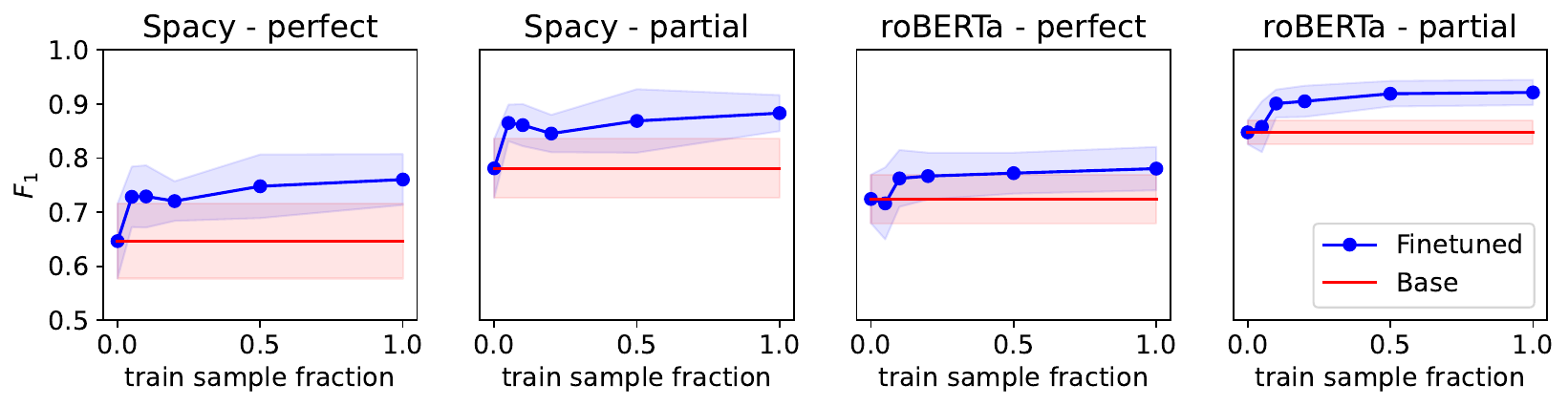}
    \end{center}
    \caption{Average $F_1$ score and standard deviation on 10 folds cross-validation at different sample size. $F_1$ score is computed for Spacy and roBERTa models, baseline and finetuned, and for perfect and partial matches.}
    \label{fig:sample_size_score}
    \Description[F1 score increases with sample size, with rise in
    performance within the first 10\%]{F1 score increases with sample size, with rise in
    performance within the first 10\%. roBERTa - partial and Spacy - partial significantly benefit from the training with full sample size}
\end{figure*}

Figure \ref{fig:sample_size_score} shows the improvement in performance as the amount of training data is increased.
We find a rise in performance within the first 10\% (about 50 documents), which gradually increases especially for the partial matches, which is true for both models, and for both testing using perfect match and partial match.
roBERTa benefits especially from the new data, achieving F1 score above 0.9 with additional data (for partial scoring).

Next, we examine the disagreements in the extracted toponyms by the different algorithms. 
In Table \ref{tab:roberta_base_vs_tuned3} we show several examples of a text fragment and the associated toponyms extracted by the two algorithms: in their baseline form and finetuned. 
The first example talks about a boat (``VOS vessel'') named Theia, which is extracted as a location by the roBERTa baseline model, but is ignored by the finetuned one (Spacy performs correctly even as a baseline).
The second example has the phrase ``Syrian Government'', which is an associative toponym that we do not select for the present task. The tuned roBERTa model learn to ignore such cases. 
Third example is similar, but we find that the training benefits Spacy, such that the associative toponym is not selected.
In the fourth example the location string is mixed with numeric characters.
Whereas the baseline roBERTa model is not able to identify it, Spacy identifies the word, but with additional numerical characters. 
Training improves roBERTa, but makes Spacy ignore the string (note that a combined model would still find the toponym).
The fifth example shows how the trained model (in this case roBERTa) is capable of recognising long lists of noun modifiers associated with ``districts''. 
Finally, the last example shows how the trained models are able to identify location at a very granular level, in this case an hospital building. 
Although not specifically focusing on particular kinds of toponyms, we find that smaller locations such as hospitals, airports, refugee camps, and others may especially necessitate training data to be caught by the named entity extractors.

\begin{table*}[!htb]
    \centering
    \caption{Sample of toponyms with their context predicted by baseline and finetuned models. Correct annotation in bold.}
    \label{tab:roberta_base_vs_tuned3}
    \begin{tabular}{p{3.5cm} p{3cm} p{3cm} p{3cm}}
        \toprule
        roBERTa$_{tuned}$ & roBERTa$_{baseline}$ & Spacy$_{tuned}$ & Spacy$_{baseline}$ \\
        \midrule
        \multicolumn{4}{p{12cm}} { \emph{1. two VOS vessels. Two will be located on board the VOS Theia, two will be located at Aden port, and one will  ...} }\\
        
        None \checkmark \rule{0pt}{2.8ex} & VOS Theia & None \checkmark & None \checkmark\\
        \hline
        \multicolumn{4}{p{12cm}} { \emph{2. ... and volatile in June, with tensions between the Syrian Government and ‘reconciled’ non-state armed groups reported ...  \rule{0pt}{2.8ex} } }\\
        None\checkmark \rule{0pt}{2.8ex} & Syrian Government & None \checkmark & None \checkmark \\
        \hline
        \multicolumn{4}{p{12cm}} {\emph{3. The US Government Congratulates Buhari in Spite of Violent and Corrupt Election \rule{0pt}{2.8ex} }}\\
        None\checkmark \rule{0pt}{2.8ex} & None\checkmark & None\checkmark & US \\
        \hline
        \multicolumn{4}{p{12cm}} {\emph{ 4. Plateau(5), Taraba(3), Gombe(1), Kaduna(1), \textbf{Kwara}(1), FCT(1), Benue(2), Rivers(1) Kogi(1)  \rule{0pt}{2.8ex} ...}}\\
        Kwara\checkmark \rule{0pt}{2.8ex} & None & None & Kwara(1 \\
        \hline
        \multicolumn{4}{p{12cm}} {\emph{ 5. However, clashes intensified in \textbf{At Tuhayat and Zabid districts} of Hudaydah city ... \rule{0pt}{2.8ex} }}\\
        At Tuhayat and Zabid districts \checkmark \rule{0pt}{2.8ex} & At Tuhayat & Tuhayat & None \\
        \hline
        \multicolumn{4}{p{12cm}} {\emph{6. sources report 17 dead and eight wounded, currently in treatment at \textbf{Am-Timan hospital \rule{0pt}{2.8ex} }}}\\
        Am-Timan hospital\checkmark \rule{0pt}{2.8ex} & None & Am-Timan hospital\checkmark & None \\
        
        \bottomrule
        \end{tabular}
\end{table*}

\subsection{Geocoding}

\subsubsection{Baseline Approaches}

Considering the geocoding task -- the mapping of toponyms to a unique set of geo-coordinates, and potentially to other meta-data associated with a location -- the literature is sparse in this area. 
We consider two approaches to geocoding from the previous literature. 
Buscaldi \& Magnini \cite{buscaldi2010grounding} propose an approach based on an iterative resolution of references by favoring ``unambiguous'' ones, with an assumption that the references appearing at greater frequency (such as countries) are less ambiguous. 
Intuitively, for each candidate geolocation for a toponym, the algorithm considers resolved references in the context around the candidate, and computes inverse distance to the unambiguous geolocations. 
The candidate geolocation having the shortest distance to the unambiguous known references in the text is chosen as the best.
Following the findings of the paper, we do not implement the text distance between the toponym and references in the text.
We considered as unambiguous toponyms which had only one GeoNames match and had to be in one of the countries mentioned in the document, as well as the countries themselves. 
The second approach by Chen, Vasardani, and Winter \cite{chen2018disambiguating} % https://arxiv.org/pdf/1808.05946.pdf
also uses the contextual information around a toponym.
It involves the clustering of the candidate geolocations for all toponyms found in a document, finding the best clustering radius, and choosing the candidate geolocations from the biggest cluster. 
The model assumes that the toponyms found in a document are located near each other.
It further does not provide a tie-breaking mechanism for choosing the best geolocation within the selected cluster (so we break ties randomly).

\subsubsection{FeatureRank}

As an alternative to the general-purpose geocoding algorithms, in this work we propose a custom feature-based geocoding method tailored for the humanitarian domain. 
This method, FeatureRank, considers geopolitical and population features of candidate locations extracted from GeoNames, as well as a document-wide distribution of the candidate locations for all toponyms, thus capturing both local and global knowledge.

First, the toponyms are preprocessed as in the Section \ref{annotation:geocoding}. 
Recall that the preprocessing involves several steps in cleaning and splitting toponyms, searching GeoNames database using string matching and search engine, and filtering the candidate geolocation set using a set of codes.
This results in a set of candidate geolocations for each toponym in a document.
A variant of our algorithm SearchFeatureRank also uses a search engine to query for candidate locations (instead of substring match).
Similar to the baseline models, we summarize the distribution of the geolocation candidates in the documents by computing a document-wide country distribution.
To compute this distribution, for each toponym in the document we first create a country distribution of its geolocation candidates, and finally average them for a document-wide distribution.
Note that we are assuming the document is largely concerning one country, which we find to be the case for most of the documents in our dataset, but is not necessarily true for other settings.
We then geocode each toponym.

To do this, we compute a set of features for each geolocation candidate for a toponym:

\begin{itemize}
    \item \texttt{IsCapital}: it is 1 if the GeoNames feature code is PPLC, and only if the candidate comes from the string match.
    \item \texttt{IsCountry}: it is 1 if the GeoNames feature code is PCLI or PCLS, and only if the candidate comes from the string match.
    \item \texttt{AdminLevel}: the numeric level (1-5) of the administrative unit (ADM or PPL) in the GeoNames feature code; 0 for capital cities or countries; 6 for geolocations without a level.
    \item \texttt{IsCity}: it is 1 if the GeoNames feature code begins with PPL (some populated location).
    \item \texttt{Population}: population provided by GeoNames.
    \item \texttt{DocCountryMatch}: the probability of the country of the candidate geolocation in the document country distribution. 
\end{itemize}

Finally, the candidates are ranked using the following order: \texttt{IsCapital} descending, \texttt{IsCountry} descending, \texttt{DocCoun-} \texttt{tryMatch} descending, \texttt{AdminLevel} ascending, \texttt{IsCity} descending, \texttt{Population} descending.
The algorithm selects the best results according to the ranking strategy, therefore a candidate is always chosen even if none of the candidates matches the correct location. 
A threshold rule has proven effective to filter out unlikely solutions: if \texttt{AdminLevel} $>$ 5 and the candidate Country is not the country with highest \texttt{DocCountryMatch}, the location is discarded.

\subsubsection{Evaluation}

We evaluate these algorithms on the Geocoding dataset described above. 
In particular, we consider several metrics:

\begin{itemize}
    \item For correct prediction, we consider two cases:
    \begin{itemize}
        \item True Positive: match exists and correctly predicted
        \item True Negative: match does not exist and is correctly ignored
        \item Correct: sum of the two above
    \end{itemize}
    \item For incorrect prediction, we consider three cases:
    \begin{itemize}
        \item False Positive (new selection): match does not exist, but some value is wrongly predicted
        \item False Negative: match exists but is not selected (i.e excluded by threshold)
        \item False Positive (wrong selection): match exists but wrong candidate is selected
        \item Incorrect: sum of the three above
    \end{itemize}
    \item For False Positive (wrong selection) we compute the average distance of the incorrect guess to the correct one.
    \item For the toponyms which were tagged as non-locations, how many were identified as locations.
\end{itemize}

We can see from Table \ref{tab:geocoding_performance2} that an approach based on features is better suited than other methods based on geographical distances. 
Both FeatureRank and SearchFeatureRank reach similar performances, with correct prediction in 79\% and 80\% of the cases, respectively. The only notable difference is in the median distance error, which decreases in the search engine version.
Note that during the geocoding annotation process, it was possible to indicate whether a toponym is correct, and in 18 cases we found that the toponym was selected incorrectly during the first round of annotations. 
Using this information, we are able to test the performance of our models on such cases, and we find that FeatureRank  correctly does not match these cases to anything 50\% of the time (compared to 0\% by BM and 22\% by Chen baselines).

\begin{table*}[!htb]
    \centering
    \caption{proportion of correct and incorrect matches, and median distance from prediction to ground truth for wrong matches for geocoding methods. We propose FeatureRank and SearchFeatureRank. }
    \label{tab:geocoding_performance2}
    \begin{tabular}{l|rrrr}
        \toprule
         & BM & Chen & FeatureRank & SearchFeatureRank\\
        \midrule
        
        Locations (count) & 543 & 543 & 543 & 543\\
        \midrule
        True Positive &0.58&0.21&0.71&0.72\\
        True Negative &0.00&0.01&0.08&0.08\\
        Correct       &0.58 &0.22 & 0.79 &0.80\\
        \midrule
        False Positive (new selection)   &0.13&0.12&0.05&0.05\\
        False Negative                   &0.00&0.03&0.04&0.03\\
        False Positive (wrong selection) &0.29&0.63&0.12&0.12\\
        Incorrect                        &0.42&0.78&0.21&0.20\\
        \midrule
        Median dist. (Km) & 219.36&2610.94&83.11&61.56\\
        \midrule
        Non-locations (count) & 18 & 18& 18& 18\\
        Correct &0.00&0.22&0.50&0.44\\
        \bottomrule
    \end{tabular}
\end{table*}

When we consider the 29 locations known to be in US \& Europe separately from all others, we find that our algorithm attains 100\% accuracy for locations in US \& Europe and 84\% for those outside (and for those toponyms which are locations, but which are not in GeoNames, the algorithm performs the worst, at 48\% accuracy). 
However, in terms of guessed locations, only 69\% of the locations our algorithm guesses to be in US \& Europe are correct, compared to 80\% of those guessed in other locales.
In particular, our algorithm works well for locations in \emph{Libya}, \emph{Colombia}, \emph{Afghanistan} and \emph{Ecuador}, where 100\% of locations are correctly predicted. 
We note that the worst performing countries are \emph{Panama}, \emph{Mexico} and \emph{Paraguay}, for which the algorithm never predicts correctly (however, we also do not have any projects about \emph{Panama} and \emph{Paraguay} in the training data, though locations from those countries could be mentioned in other projects). 
We hypothesize it is due to the similar naming and language conventions among these countries.

\section{Application study}

\begin{figure*}[!htbp]
    \begin{center}
        \includegraphics[width=12cm]{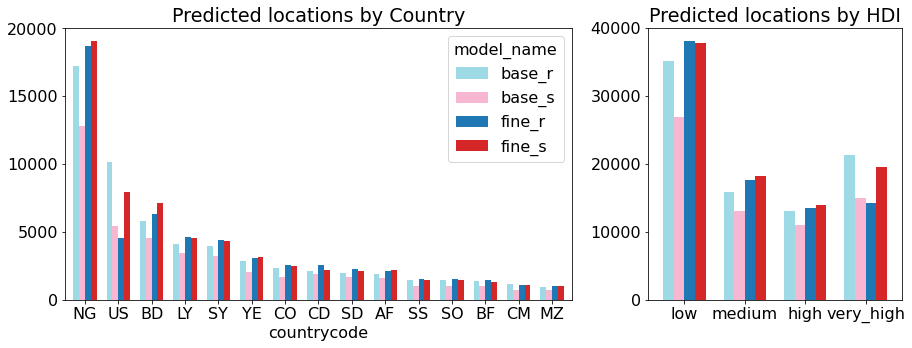}
    \end{center}
    \caption{Number of locations detected in HumSet by baseline Spacy (s) and roBERTa (r) models, as well as the tuned ones by country (left, top 15 countries shown), and by the country's human development index (right).}
    \label{fig:humset_locs_by_country}
    \Description[Overall, roBERTa trained model finds fewer locations in US and countries with very-high HDI]{Overall, roBERTa trained model finds fewer locations in US and countries with very-high HDI. On the other hand, it find more locations in countries from low to high HDI. Spacy increases the number of locations found across all the HDI levels.}
\end{figure*}

Finally, we apply the tuned toponym extraction and the custom FeatureRank geocoding algorithm to the HumSet dataset.
Although we do not have the ground truth for these labels, we examine potential biases in the locations that the baseline models find, compared to our tuned ones.
In total, we annotate 6733 documents, extracting \num{13967} distinct locations.
Figure \ref{fig:humset_locs_by_country} shows the number of locations identified by the two baseline models, and two tuned models, by country (left) and by country grouped by their human development index (right).
We find that, generally, the number of matched locations increases for both models. 
However, for roBERTa, the matches in US decrease dramatically, which is also reflected in the lower matches in countries with very high HDI.
Note that this effect was not intended by the tuning of the algorithm, and is an interesting side effect. We can also notice that the number of locations detected increases even in countries that were not in the training set, like Democratic Republic Of the Congo (CD) and Mozambique (MZ).

\begin{figure*}[!htbp]
    \begin{center}
        \includegraphics[width=11cm]{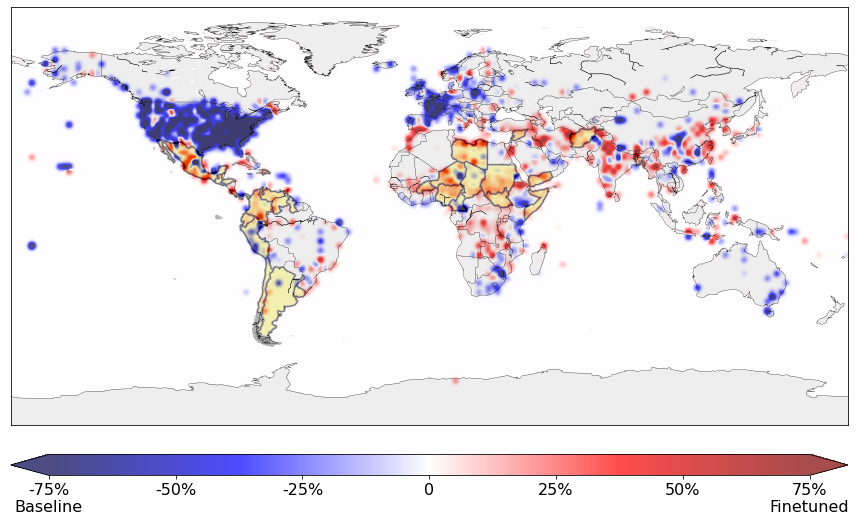}
    \end{center}
    \caption{Map of the difference in locations identified by roBERTa: in red more by the tuned model, and in blue more by the baseline model. Countries in yellow are those for which HumSet contains projects. }
    \label{fig:map}
    \Description[Baseline models find more locations in Western countries. Finetuned models finds more locations in countries included in HumSet]{Baseline models find more locations in Western countries, in particular US, European countries and Australia. Finetuned models finds more locations in countries included in HumSet}
\end{figure*}

Figure \ref{fig:map} shows a map with areas highlighted in blue where the baseline roBERTa model finds more locations, and highlighted in red where the tuned roBERTa model finds more. 
We observe that the baseline model favors United States and Europe, and Australia -- countries with high and very high development index. In fact, the tuned model predicts 55\% fewer matches in US.
On the other hand, the countries highlighted in yellow -- those for which there are projects in the HumSet data -- show an increase in matches, especially Trinidad and Tobago has more than twice as many matches (from 92 by baseline to 191 by tuned), Domenican Republic increases by 17\%, Bahamas by 16\%, Sudan 14\%, and Lybia by 12\%.
However, even countries which do not have projects in the HumSet documents have more matches, including Morocco at 102\% more, South Korea 90\%, North Korea 81\%, and Myanmar 64\%.
As we do not know the ground truth of these matches, more work needs to be done to understand whether this is due to increase in accuracy, or whether this is an evidence of a new geographic bias.
If these new matches are indeed correct, this would point to a generalizability of these models beyond the countries of the tuned data.

\section{Discussion}

As we attempted to improve the performance of state-of-the-art NER for geolocation of entities in humanitarian texts, we discovered that training data from the humanitarian domain has not only improved accuracy of the tools, but also possibly helped alleviate the geographic bias of these tools toward the Western countries.
Our findings during the application of the baseline and tuned algorithm to the HumSet data suggest that data extraction tools should be tested for bias in a systematic way, as proposed by algorithmic auditing proponents \cite{vecchione2021algorithmic}.
The social justice concerns auditors of economic and governmental systems extend to the use of AI tools in regards to possibly vulnerable populations, such as those needing humanitarian assistance. 
Similar concerns have been raised in the use of AI for medical diagnosis \cite{seyyed2021underdiagnosis}, hiring \cite{raghavan2020mitigating}, and poverty mapping \cite{beiro2022fairness, sartirano2023strengths}.
Auditing is especially necessary for black-box algorithms wherein coverage and completeness are impossible to ascertain, unlike for algorithms that use gazetteers or databases.

In humanitarian context, geographic accuracy is imperative for the accurate assessment of needs and their development over time.
During the development of the annotated resources, we encountered several peculiarities of geolocations present in humanitarian texts.
First, we often encounter geo-relevant organization names, such as ``Syrian Arab Red Crescent'' or ``European Commissioner for Crisis Management'' that indicate their affiliation or source, but not necessarily the location of their current involvement. 
We observed that this often would decrease the performance of distance-based algorithms that average locations of found entity candidates.
A specialized module to handle such information may benefit the mapping of the entities involved in a project.
Second, locations often have qualifiers such as ``western Yemen'' or ``north of Baghdad'', necessitating for additional semantics in location representation that could capture both the directionality and uncertainty of the location.
Third, rapid developments on the ground, such as the building or dismantling refugee camps, should be added to the geographic resources used for humanitarian aid on an ongoing basis.

More research is needed to improve these tools and evaluate them on a larger set of documents. 
Further, thus far only documents in English language were considered (although many humanitarian organizations publish their reports in English for international consumption). 
This paper does not use any personal data, instead it uses only published reports and news articles, available under the Apache 2.0 license. 
We hope the annotated resources and guidelines provided in this paper will spur further work on tuning NER tools to handle data around vulnerable populations.

\begin{acks}
    The authors gratefully acknowledge the support from the Lagrange Project of the Institute for Scientific Interchange Foundation (ISI Foundation) funded by Fondazione Cassa di Risparmio di Torino (Fondazione CRT).
\end{acks}
\bibliographystyle{ACM-Reference-Format}
\bibliography{bibliography}

\end{document}